\documentclass[10pt,twocolumn,letterpaper]{article}

\usepackage{cvpr}
\usepackage{times}
\usepackage{epsfig}
\usepackage{graphicx}
\usepackage{amsmath}
\usepackage{amssymb}
\usepackage{xcolor}
\usepackage{dsfont}

\usepackage[utf8]{inputenc} 
\usepackage[T1]{fontenc}    
\usepackage{url}            
\usepackage{booktabs}       
\usepackage{amsfonts}       
\usepackage{nicefrac}       
\usepackage{microtype}      
\usepackage{graphicx}
\usepackage{mathabx}
\usepackage{longtable}
\usepackage{amsthm}   
\usepackage{stackrel}  

\usepackage{bm} 
\usepackage{subfig}
\usepackage{float}  
\usepackage{multirow}
\usepackage{xcolor,colortbl} 
\usepackage{color}  
\usepackage[linesnumbered,ruled]{algorithm2e}  
\usepackage[skip=2pt]{caption}  

\DeclareCaptionFormat{myformat}{\small\selectfont#1#2#3}
\usepackage[pagebackref=true,breaklinks=true,letterpaper=true,colorlinks,bookmarks=false]{hyperref}

\cvprfinalcopy 

\def\cvprPaperID{9690}

\providecommand{\realnum}{\mathbb{R}}
\providecommand{\naturalnum}{\mathbb{N}}
\providecommand{\bmcal}[1]{\bm{\mathcal{#1}}}

 \providecommand{\matnot}[1]{_{[{#1}]}}  
 \providecommand{\invar}{z}  
\providecommand{\binvar}{\bm{\invar}}  
\providecommand{\outvar}{x}  
\providecommand{\boutvar}{\bm{\outvar}}  
\providecommand{\citep}{\cite} 
\providecommand{\citet}{\cite}

\newcommand{\modelname}{ProdPoly}
\newcommand{\modelone}{CCP}
\newcommand{\modeltwo}{NCP}
\newcommand{\modelthree}{NCP-Skip}
\newcommand{\resnet}{ResNet}
\newcommand{\modelres}{Prodpoly-\resnet}

\ifcvprfinal\fi
\begin{document}

\title{$\Pi-$nets: Deep Polynomial Neural Networks}

\author{
Grigorios G. Chrysos$^{1}$, \quad Stylianos Moschoglou$^{1,2}$, \quad Giorgos Bouritsas$^{1}$, \quad Yannis Panagakis$^{3}$, \\ \quad Jiankang Deng$^{1,2}$, \quad Stefanos Zafeiriou$^{1,2}$ \\ 
{\textsuperscript{1} Department of Computing, Imperial College London, UK}\\
{\textsuperscript{2} Facesoft.io}\\
{\textsuperscript{3} Department of Informatics and Telecommunications , University of Athens, GR}\\
{\texttt{\{[first letter].[surname]\}@imperial.ac.uk}}
}
\maketitle

\begin{abstract}
  Deep Convolutional Neural Networks (DCNNs) is currently the method of choice both for generative, as well as for discriminative learning in computer vision and machine learning. The success of DCNNs can be attributed to the careful selection of their building blocks (e.g., residual blocks, rectifiers, sophisticated normalization schemes, to mention but a few). In this paper, we propose $\Pi$-Nets, a new class of DCNNs. $\Pi$-Nets are polynomial neural networks, i.e., the output is a high-order polynomial of the input. $\Pi$-Nets can be implemented using special kind of skip connections and their parameters can be represented via high-order tensors. We empirically demonstrate that $\Pi$-Nets have better representation power than standard DCNNs and they even produce good results \textbf{without the use of non-linear activation functions} in a large battery of tasks and signals, i.e., images, graphs, and audio. When used in conjunction with activation functions, $\Pi$-Nets produce state-of-the-art results in challenging tasks, such as image generation. Lastly, our framework elucidates why recent generative models, such as StyleGAN, improve upon their predecessors, e.g., ProGAN.   
\end{abstract}

\section{Introduction}
\label{sec:prodpoly_introduction}

\begin{figure}[!t]
    \centering
    \includegraphics[width=1\linewidth]{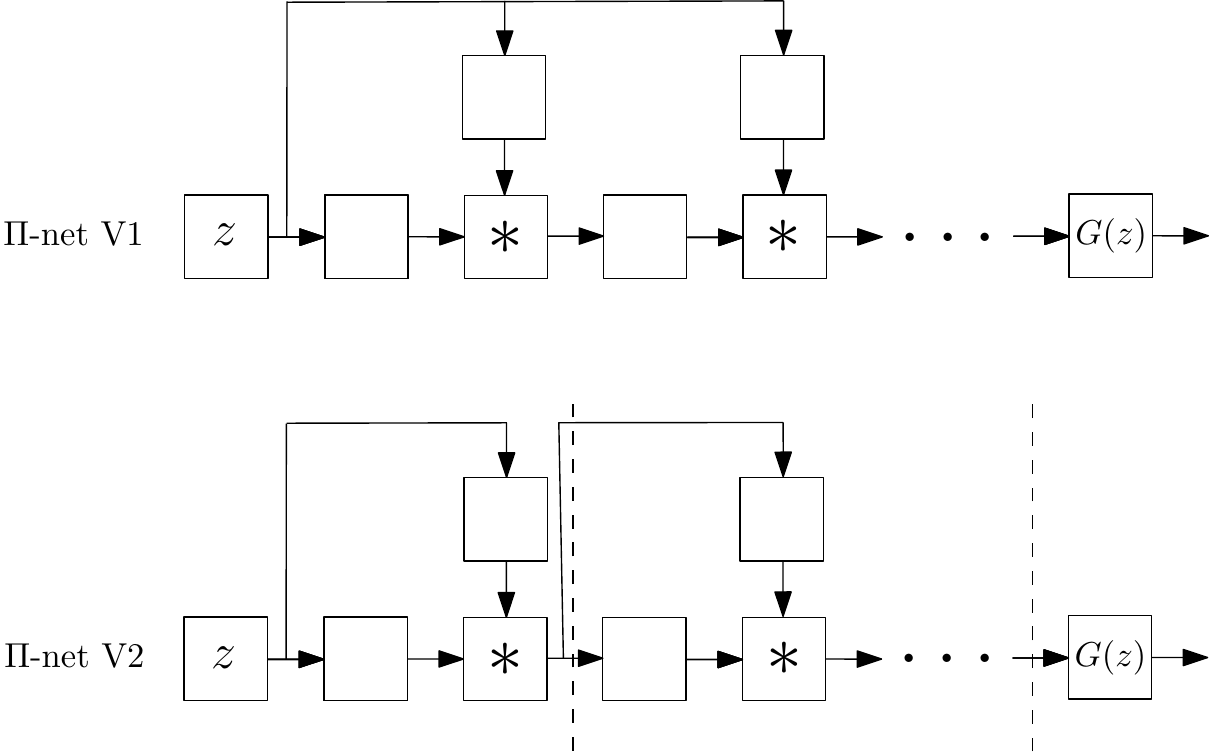}
\caption{In this paper we introduce a class of networks called $\Pi-$nets, where the output is a polynomial of the input. The input in this case, $\binvar$, can be either the latent space of Generative Adversarial Network for a generative task or an image in the case of a discriminative task. Our polynomial networks can be easily implemented using a special kind of skip connections.}
\label{fig:prodpoly_model_intro_schematic}
\end{figure}

Representation learning via the use of (deep) multi-layered  non-linear models has revolutionised the field of computer vision the past decade~\cite{krizhevsky2012imagenet, he2016deep}. Deep Convolutional Neural Networks (DCNNs)~\cite{lecun1998gradient, krizhevsky2012imagenet} have been the dominant class of models. Typically, a DCNN is a sequence of layers where the output of each layer is fed first to a convolutional operator (i.e., a set of shared weights applied via the convolution operator) and then to a non-linear activation function. Skip connections between various layers allow deeper representations and improve the gradient flow while training the network~\cite{he2016deep, srivastava2015highway}. 

In the aforementioned case, if the non-linear activation functions are removed, the output of a DCNN degenerates to a linear function of the input. In this paper, we propose a new class of DCNNs, which we coin $\Pi-$nets, where the output is a polynomial function of the input. We design $\Pi-$nets for generative tasks (e.g., where the input is a small dimensional noise vector) as well as for discriminative tasks (e.g., where the input is an image and the output is a vector with dimensions equal to the number of labels). We demonstrate that these networks can produce good results without the use of non-linear activation functions. Furthermore, our extensive experiments show, empirically, that $\Pi-$nets can consistently improve the performance, in both generative and discriminative tasks, using, in many cases, significantly fewer parameters. 

DCNNs have been used in computer vision for over 30 years~\cite{lecun1998gradient, schmidhuber2015deep}. Arguably, what brought DCNNs again in mainstream research was the remarkable results achieved by the so-called AlexNet in the ImageNet challenge~\cite{krizhevsky2012imagenet}. Even though it is only seven years from this pioneering effort the field has witnessed dramatic improvement in all data-dependent tasks, such as  object detection~\cite{huang2017densely} and image generation~\cite{miyato2018spectral, gulrajani2017improved}, just to name a few examples. The improvement is mainly attributed to carefully selected units in the architectural pipeline of DCNNs, such as blocks with skip connections~\cite{he2016deep}, sophisticated normalization schemes (e.g., batch normalisation~\cite{ioffe2015batch}), as well as the use of efficient gradient-based optimization techniques~\cite{kingma2014adam}.

Parallel to the development of DCNN architectures for discriminative tasks, such as classification, the notion of Generative Adversarial Networks (GANs) was introduced for training generative models.
GANs became instantly a popular line of research but it was only after the careful design of DCNN pipelines and training strategies that GANs were able to produce realistic images~\cite{karras2018style, brock2019large}. ProGAN~\cite{karras2017progressive} was the first architecture to synthesize realistic facial images by a DCNN. StyleGAN~\cite{karras2018style} is a follow-up work that improved ProGAN. The main addition of StyleGAN was a type of skip connections, called ADAIN~\cite{huang2017arbitrary}, which allowed the latent representation to be infused in all different layers of the generator. Similar infusions were introduced in \cite{park2019semantic} for conditional image generation. 

Our work is motivated by the improvement of StyleGAN over ProGAN by such a simple infusion layer and the need to provide an explanation\footnote{The authors argued that this infusion layer is a kind of a style that allows a coarser to finer manipulation of the generation process. We, instead, attribute this to gradually increasing the power of the polynomial.}. We show that such infusion layers create a special non-linear structure, i.e., a higher-order polynomial, which empirically improves the representation power of DCNNs. We show that this infusion layer can be generalized (e.g. see Fig.~\ref{fig:prodpoly_model_intro_schematic}) and applied in various ways in both generative, as well as discriminative architectures. In particular, the paper bears the following contributions:
\begin{itemize}
    \item
    We propose a new family of neural networks (called $\Pi-$nets) where the output is a high-order polynomial of the input. To avoid the combinatorial explosion in the number of parameters of polynomial activation functions~\cite{kileel2019expressive} our $\Pi-$nets use a special kind of skip connections to implement the polynomial expansion (please see Fig. \ref{fig:prodpoly_model_intro_schematic} for a brief schematic representation). We theoretically demonstrate that these kind of skip connections relate to special forms of tensor decompositions.
    \item We show how the proposed architectures can be applied in generative models such as GANs, as well as discriminative networks. We showcase that the resulting architectures can be used to learn high-dimensional distributions without non-linear activation functions. 
    \item We convert state-of-the-art baselines using the proposed $\Pi-$nets and show how they can largely improve the expressivity of the baseline. We demonstrate it conclusively in a battery of tasks (i.e., generation and classification). Finally, we demonstrate that our architectures are applicable to many different signals such as images, meshes, and audio.
\end{itemize}

 \section{Related work}
\label{sec:prodpoly_related}

\textbf{Expressivity of (deep) neural networks}: The last few years, (deep) neural networks have been applied to a wide range of applications with impressive results. The performance boost can be attributed to a host of factors including: a) the availability of massive datasets~\cite{deng2009imagenet, liu2015deep}, b) the machine learning libraries~\cite{chainer_learningsys2015, paszke2017automatic} running on massively parallel hardware, c) training improvements. The training improvements include a) optimizer improvement~\cite{kingma2014adam, reddi2019convergence}, b) augmented capacity of the network~\cite{simonyan2014very}, c) regularization tricks~\cite{glorot2010understanding, saxe2013exact, ioffe2015batch, ulyanov2016instance}. However, the paradigm for each layer remains largely unchanged for several decades: each layer is composed of a linear transformation and an element-wise activation function. Despite the variety of linear transformations~\cite{fukushima1980neocognitron, lecun1998gradient, krizhevsky2012imagenet} and activation functions~\cite{ramachandran2017searching, nair2010rectified} being used, the effort to extend this paradigm has not drawn much attention to date.

Recently, hierarchical models have exhibited stellar performance in learning expressive generative models~\cite{brock2019large, karras2018style, zhao2017learning}. For instance, the recent BigGAN \citep{brock2019large} performs a hierarchical composition through skip connections from the noise $\bm{z}$ to multiple resolutions of the generator. A similar idea emerged in StyleGAN~\citep{karras2018style}, which is an improvement over the Progressive Growing of GANs (ProGAN)~\citep{karras2017progressive}. As ProGAN, StyleGAN is a highly-engineered network that achieves compelling results on synthesized 2D images. In order to provide an explanation on the improvements of StyleGAN over ProGAN, the authors adopt arguments from the style transfer literature \citep{huang2017arbitrary}. We believe that these improvements can be better explained under the light of our proposed polynomial function approximation. Despite the hierarchical composition proposed in these works, we present an intuitive and mathematically elaborate method to achieve a more precise approximation with a polynomial expansion. We also demonstrate that such a polynomial expansion can be used in both image generation (as in \cite{karras2018style, brock2019large}), image classification, and graph representation learning. 

\textbf{Polynomial networks}: Polynomial relationships have been investigated in two specific categories of networks: a) self-organizing networks with hard-coded feature selection, b) pi-sigma networks.

The idea of learnable polynomial features can be traced back to Group Method of Data Handling (GMDH)~\cite{ivakhnenko1971polynomial}\footnote{This is often referred to as the first deep neural network~\cite{schmidhuber2015deep}.}. GMDH learns partial descriptors that capture quadratic correlations between two predefined input elements. In \cite{oh2003polynomial}, more input elements are allowed, while higher-order polynomials are used. The input to each partial descriptor is predefined (subset of the input elements), which does not allow the method to scale to high-dimensional data with complex correlations.

Shin \etal~\cite{shin1991pi} introduce the pi-sigma network, which is a neural network with a single hidden layer. Multiple affine transformations of the data are learned; a product unit multiplies all the features to obtain the output. Improvements in the pi-sigma network include regularization for training in~\cite{xiong2007training} or using multiple product units to obtain the output in~\cite{voutriaridis2003ridge}. 
The pi-sigma network is extended in sigma-pi-sigma neural network (SPSNN)~\cite{li2003sigma}. The idea of SPSNN relies on summing different pi-sigma networks to obtain each output. SPSNN also uses a predefined basis (overlapping rectangular pulses) on each pi-sigma sub-network to filter the input features. Even though such networks use polynomial features or products, they do not scale well in high-dimensional signals. In addition, their experimental evaluation is conducted only on signals with known ground-truth distributions (and with up to 3 dimensional input/output), unlike the modern generative models where only a finite number of samples from high-dimensional ground-truth distributions is available.

 \section{Method}
\label{sec:prodpoly_method}

\begin{table*}[h]
\caption{Nomenclature}
\label{tbl:prodpoly_primary_symbols}
\centering
\begin{tabular}{|c | c | c|}
\toprule
Symbol 	& Dimension(s) 		&	Definition \\
\midrule
$n, N$ 		            & $\naturalnum$		            &	Polynomial term order, total approximation order. \\
$k$ 		            & $\naturalnum$		            & Rank of the decompositions. \\
$\binvar$            & $\realnum^d$                      & Input to the polynomial approximator, i.e., generator. \\
$\bm{C}, \bm{\beta}$ 		    & $\realnum^{o\times k}, \realnum^{o}$		        &	Parameters in both decompositions. \\
$\bm{A}\matnot{n}, \bm{S}\matnot{n}, \bm{B}\matnot{n}$          &       $\realnum^{d\times k}, \realnum^{k\times k}, \realnum^{\omega\times k}$     & Matrix parameters in the hierarchical decomposition.\\
$\odot, *$          &   -       & Khatri-Rao product, Hadamard product. \\
 \hline
\end{tabular}
\end{table*}

\textbf{Notation}: Tensors are symbolized by calligraphic letters, e.g.,  $\bmcal{X}$, while matrices (vectors) are denoted by uppercase (lowercase) boldface letters e.g., $\bm{X}$, ($\bm{x}$). The \textit{mode-$m$ vector product} of $\bmcal{X}$ with a
vector $\bm{u} \in \realnum^{I_m}$ is denoted by
$\bmcal{X} \times_{m} \bm{u}$.\footnote{A detailed tensor notation is provided in the supplementary.}

We want to learn a function approximator where each element of the output $\outvar_j$, with $j\in [1, o]$, is expressed as a polynomial\footnote{The theorem of \citep{stone1948generalized} guarantees that any smooth function can be approximated by a polynomial. The approximation of multivariate functions is covered by an extension of the Weierstrass theorem, e.g. in \cite{nikol2013analysis} (pg 19).} of all the input elements $\invar_i$, with $i\in [1, d]$. That is, we want to learn a function $G: \realnum^{d} \to \realnum^{o}$ of order $N \in \naturalnum$, such that:

\begin{equation}
\begin{split}
    \outvar_j = G(\binvar)_j = \beta_j + {\bm{w}_j^{[1]}}^T\binvar + \binvar^T \bm{W}_j^{[2]}\binvar + \\
    \bmcal{W}_j^{[3]}\times_1\binvar\times_2\binvar\times_3\binvar + \cdots + \bmcal{W}_j^{[N]}\prod_{n=1}^N \times_{n} \binvar
\end{split}
\label{eq:prodpoly_starting_poly_eq_element}
\end{equation}

where $\beta_j \in \realnum$, and  $\big\{\bmcal{W}_j^{[n]} \in \realnum^{\prod_{m=1}^n\times_m d}\big\}_{n=1}^N$ are parameters for approximating the output $\outvar_j$. The correlations (of the input elements $\invar_i$) up to $N^{th}$ order emerge in \eqref{eq:prodpoly_starting_poly_eq_element}. 
A more compact expression of \eqref{eq:prodpoly_starting_poly_eq_element} is obtained by vectorizing the outputs:

\begin{equation}
    \boutvar = G(\binvar) = \sum_{n=1}^N \bigg(\bmcal{W}^{[n]} \prod_{j=2}^{n+1} \times_{j} \binvar\bigg) + \bm{\beta}
    \label{eq:prodpoly_poly_general_eq}
\end{equation}

where $\bm{\beta} \in \realnum^o$ and $\big\{\bmcal{W}^{[n]} \in  \realnum^{o\times \prod_{m=1}^{n}\times_m d}\big\}_{n=1}^N$ are the learnable parameters. This form of \eqref{eq:prodpoly_poly_general_eq} allows us to approximate any smooth function (for large $N$), however the parameters grow with $\mathcal{O}(d^N)$. 

A variety of methods, such as pruning~\cite{frankle2018lottery, han2015learning}, tensor decompositions~\cite{kolda2009tensor, sidiropoulos2017tensor}, special linear operators~\cite{ding2017c} with reduced parameters, parameter sharing/prediction~\cite{yunpeng2017sharing, denil2013predicting}, can be employed to reduce the parameters. In contrast to the heuristic approaches of pruning or prediction, we describe below two principled ways which allow an efficient implementation. The first method relies on performing an off-the-shelf tensor decomposition on \eqref{eq:prodpoly_poly_general_eq}, while the second considers the final polynomial as the product of lower-degree polynomials. 

The tensor decompositions are used in this paper to provide a theoretical understanding (i.e., what is the order of the polynomial used) of the proposed family of $\Pi$-nets. Implementation-wise the incorporation of different $\Pi$-net structures is as simple as the incorporatation of a skip-connection. Nevertheless, in  $\Pi$-net different skip connections lead to different kinds of polynomial networks.

\subsection{Single polynomial}
\label{ssec:prodpoly_single_poly}
A tensor decomposition on the parameters is a natural way to reduce the parameters and to implement \eqref{eq:prodpoly_poly_general_eq} with a neural network. Below, we demonstrate how three such decompositions result in novel architectures for a neural network training. The main symbols are summarized in Table~\ref{tbl:prodpoly_primary_symbols}, while the equivalence between the recursive relationship and the polynomial is analyzed in the supplementary.

\begin{figure}[!h]
    \centering
    \includegraphics[width=1\linewidth]{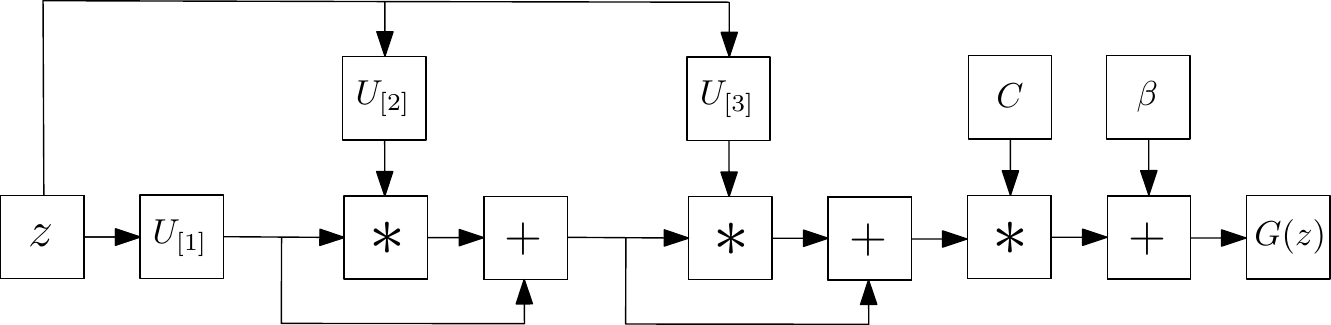}
\caption{Schematic illustration of the \modelone{} (for third order approximation). Symbol $*$ refers to the Hadamard product.}
\label{fig:prodpoly_model1_schematic}
\end{figure}

\textbf{Model 1: \modelone}: A coupled CP decomposition~\cite{kolda2009tensor} is applied on the parameter tensors. That is, each parameter tensor, i.e. $\bmcal{W}^{[n]}$ for $n\in[1, N]$, is not factorized individually, but rather a coupled factorization of the parameters is defined. The recursive relationship is:

\begin{equation}
    \boutvar_{n} = \Big(\bm{U}\matnot{n}^T \binvar \Big)* \boutvar_{n-1} + \boutvar_{n-1}
\end{equation}
for $n=2,\ldots,N$ with $\boutvar_{1} = \bm{U}\matnot{1}^T \binvar$ and $\boutvar = \bm{C}\boutvar_{N} + \bm{\beta}$. The parameters $\bm{C} \in \realnum^{o\times k}, \bm{U}\matnot{n} \in  \realnum^{d\times k}$ for $n=1,\ldots,N$ are learnable. To avoid overloading the diagram, a schematic assuming a third order expansion ($N=3$) is illustrated in Fig.~\ref{fig:prodpoly_model1_schematic}.

\textbf{Model 2: \modeltwo}: Instead of defining a flat CP decomposition, we can utilize a joint hierarchical decomposition on the polynomial parameters. A nested coupled CP decomposition (\modeltwo), which results in the following recursive relationship for $N^{th}$ order approximation is defined:
\begin{equation}
    \boutvar_{n} = \Big(\bm{A}\matnot{n}^T\binvar\Big) * \Big(\bm{S}\matnot{n}^T \boutvar_{n-1} + \bm{B}\matnot{n}^T\bm{b}\matnot{n}\Big)
    \label{eq:prodpoly_model2}
\end{equation}
for $n=2,\ldots,N$ with $\boutvar_{1} = \Big(\bm{A}\matnot{n}^T\binvar\Big) * \Big( \bm{B}\matnot{n}^T\bm{b}\matnot{n} \Big)$ and $\boutvar = \bm{C}\boutvar_{N} + \bm{\beta}$. The parameters $\bm{C} \in  \realnum^{o\times k}, \bm{A}\matnot{n} \in  \realnum^{d\times k}, \bm{S}\matnot{n} \in  \realnum^{k\times k}, \bm{B}\matnot{n} \in  \realnum^{\omega\times k}$, $\bm{b}\matnot{n} \in  \realnum^{\omega}$ for $n=1,\ldots,N$, are learnable. The explanation of each variable is elaborated in the supplementary, where the decomposition is derived.

\begin{figure}[!h]
    \centering
    \includegraphics[width=1\linewidth]{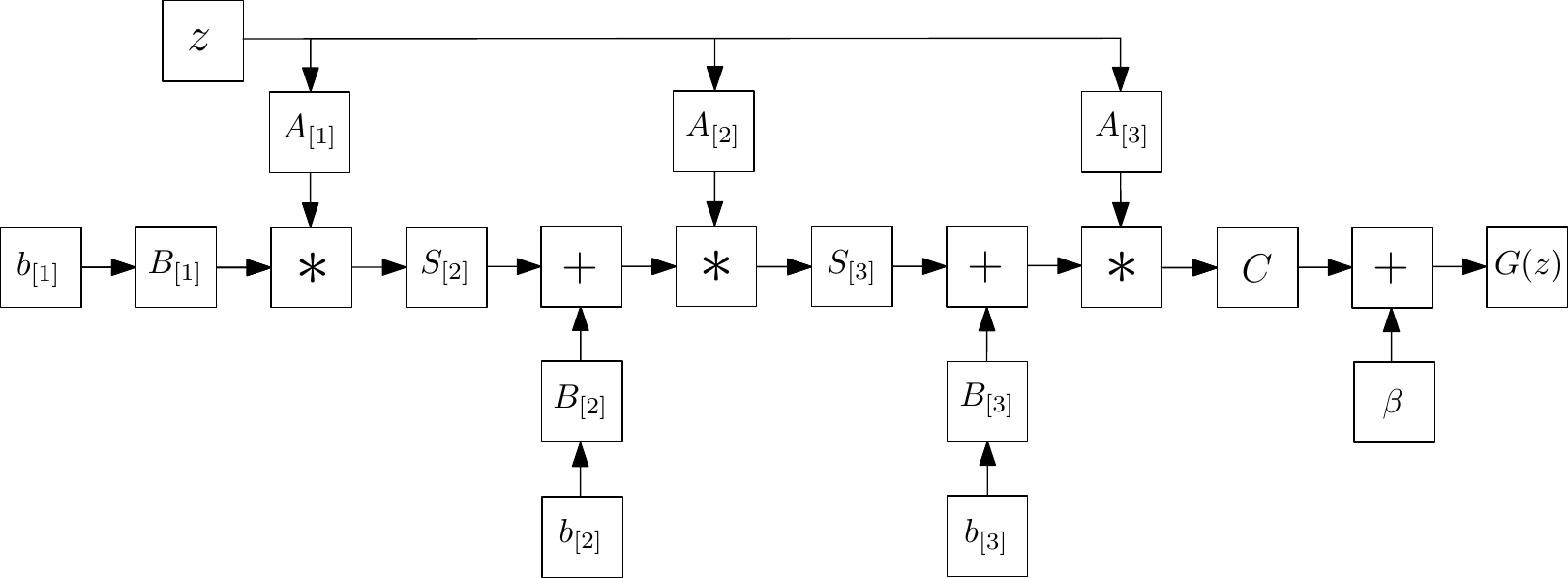}
\caption{Schematic illustration of the \modeltwo{} (for third order approximation). Symbol $*$ refers to the Hadamard product.}
\label{fig:prodpoly_model2_schematic}
\end{figure}

\textbf{Model 3: \modelthree}: The expressiveness of \modeltwo{} can be further extended using a skip connection (motivated by \modelone). The new model uses a nested coupled decomposition and has the following recursive expression:

\begin{equation}
    \boutvar_{n} = \Big(\bm{A}\matnot{n}^T\binvar\Big) * \Big(\bm{S}\matnot{n}^T \boutvar_{n-1} + \bm{B}\matnot{n}^T\bm{b}\matnot{n}\Big) +  \boutvar_{n-1}
\end{equation}

for $n=2,\ldots,N$ with $\boutvar_{1} = \Big(\bm{A}\matnot{n}^T\binvar\Big) * \Big( \bm{B}\matnot{n}^T\bm{b}\matnot{n} \Big)$ and $\boutvar = \bm{C}\boutvar_{N} + \bm{\beta}$. The learnable parameters are the same as in \modeltwo, however the difference in the recursive form results in a different polynomial expansion and thus architecture.

\begin{figure}[!h]
    \centering
    \includegraphics[width=1\linewidth]{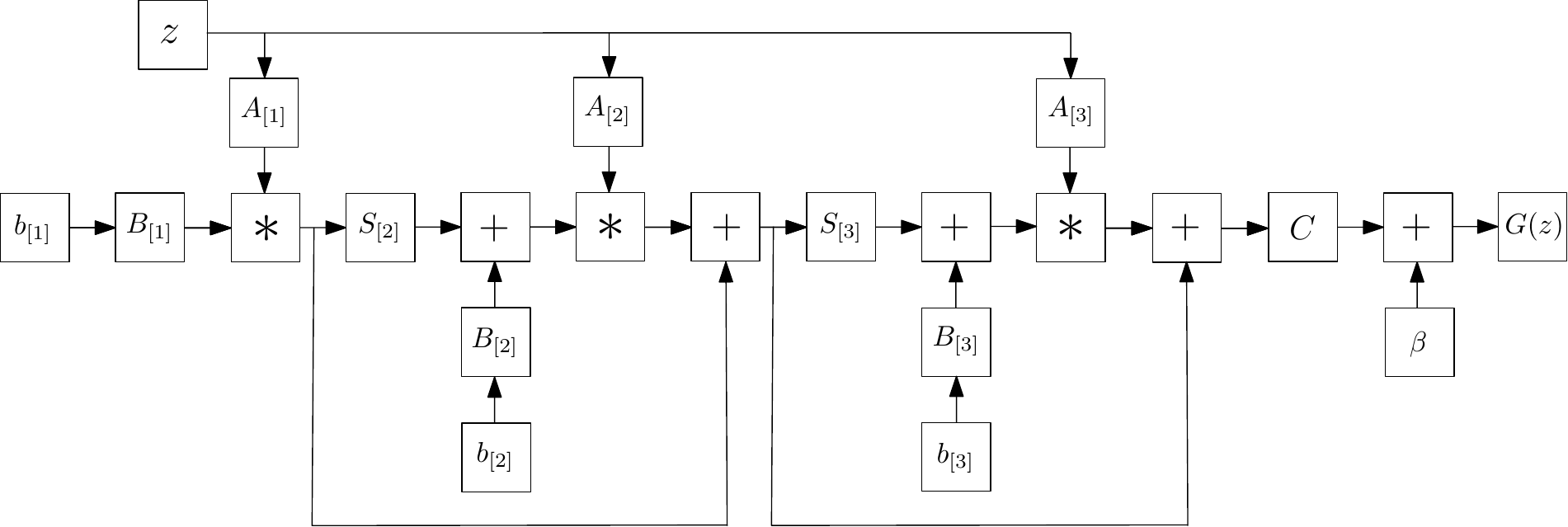}
\caption{Schematic illustration of the \modelthree{} (for third order approximation). The difference from Fig.~\ref{fig:prodpoly_model2_schematic} is the skip connections added in this model.}
\label{fig:prodpoly_model3_schematic}
\end{figure}

\textbf{Comparison between the models}: All three models are based on a polynomial expansion, however their recursive forms and employed decompositions differ. The \modelone{} has a simpler expression, however the \modeltwo{} and the \modelthree{} relate to standard architectures using hierarchical composition that have recently yielded promising results in both generative and discriminative tasks. In the remainder of the paper, for comparison purposes we use the \modeltwo{} by default for the image generation and \modelthree{} for the image classification. In our preliminary experiments, \modelone{} and \modeltwo{} share a similar performance based on the setting of Sec.~\ref{sec:prodpoly_linear_experiments}. In all cases, to mitigate stability issues that might emerge during training, we employ certain normalization schemes that constrain the magnitude of the gradients. An in-depth theoretical analysis of the architectures is deferred to a future version of our work.

\subsection{Product of polynomials}
\label{ssec:prodpoly_product_poly}
\begin{figure}[!h]
    \centering
    \includegraphics[width=1\linewidth]{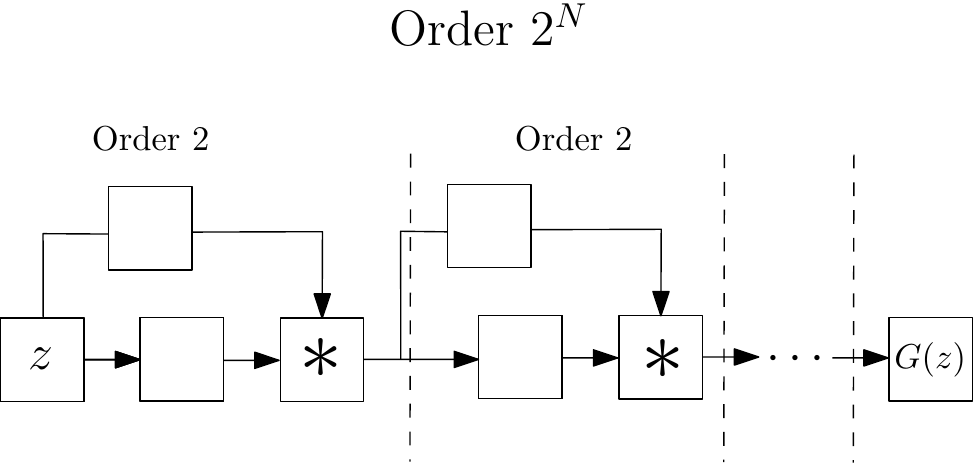}
\caption{Abstract illustration of the \modelname. The input variable $\binvar$ on the left is the input to a $2^{nd}$ order expansion; the output of this is used as the input for the next polynomial (also with a $2^{nd}$ order expansion) and so on. If we use $N$ such polynomials, the final output $G(\binvar)$ expresses a $2^N$ order expansion. In addition to the high order of approximation, the benefit of using the product of polynomials is that the model is flexible, in the sense that each polynomial can be implemented as a different decomposition of Sec.~\ref{ssec:prodpoly_single_poly}.}
\label{fig:prodpoly_prod_schematic}
\end{figure}

Instead of using a single polynomial, we express the function approximation as a product of polynomials. The product is implemented as successive polynomials where the output of the $i^{th}$ polynomial is used as the input for the $(i+1)^{th}$ polynomial. The concept is visually depicted in Fig.~\ref{fig:prodpoly_prod_schematic}; each polynomial expresses a second order expansion. Stacking $N$ such polynomials results in an overall order of $2^N$. Trivially, if the approximation of each polynomial is $B$ and we stack $N$ such polynomials, the total order is $B^N$. The product does not necessarily demand the same order in each polynomial, the expressivity and the expansion order of each polynomial can be different and dependent on the task, e.g. for generative tasks that the resolution increases progressively, the expansion order could increase in the last polynomials. In all cases, the final order will be the product of each polynomial.

There are two main benefits of the product over the single polynomial: a) it allows using different decompositions (e.g. like in Sec.~\ref{ssec:prodpoly_single_poly}) and expressive power for each polynomial; b) it requires much less parameters for achieving the same order of approximation. Given the benefits of the product of polynomials, we assume below that a product of polynomials is used, unless explicitly mentioned otherwise. The respective model of product polynomials is called \modelname.

\subsection{Task-dependent input/output}
\label{ssec:prodpoly_method_details}

The aforementioned polynomials are a function $\boutvar = G(\binvar)$, where the input/output are task-dependent. For a generative task, e.g. learning a decoder, the input $\binvar$ is typically some low-dimensional noise, while the output is a high-dimensional signal, e.g. an image. For a discriminative task the input $\binvar$ is an image; for a domain adaptation task the signal $\binvar$ denotes the source domain and $\boutvar$ the target domain.

 \section{Proof of concept}
\label{sec:prodpoly_linear_experiments}

In this Section, we conduct motivational experiments in both generative and discriminative tasks to demonstrate the expressivity of $\Pi-$nets. Specifically, the networks are implemented \textbf{without activation functions}, i.e. only linear operations (e.g. convolutions) and Hadamard products are used. In this setting, the output is linear or multi-linear with respect to the parameters.

\subsection{Linear generation}
\label{ssec:prodpoly_linear_generation}

One of the most popular generative models is Generative Adversarial Nets (GAN)~\cite{goodfellow2014generative}. We design a GAN, where the generator is implemented as a product of polynomials (using the \modeltwo\ decomposition), while the discriminator of \cite{miyato2018spectral} is used. 
No activation functions are used in the generator, but a single hyperbolic tangent ($tanh$) in the image space \footnote{\label{ft:prodpoly_supplementary}Additional details are deferred to the supplementary material.}. 

\begin{figure}
\centering
    \centering
    \includegraphics[width=1\linewidth]{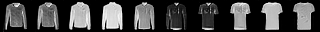} \\
    \includegraphics[width=1\linewidth]{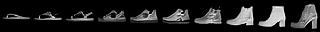} \\
    \includegraphics[width=1\linewidth]{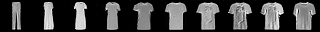} \\
    \includegraphics[width=1\linewidth]{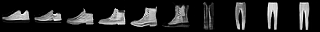} \\
    \includegraphics[width=1\linewidth]{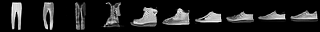} \\
    \includegraphics[width=1\linewidth]{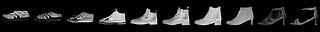}
\caption{Linear interpolation in the latent space of \modelname{} (when trained on fashion images~\cite{xiao2017fashion}). Note that the generator does not include any activation functions in between the linear blocks (Sec.~\ref{ssec:prodpoly_linear_generation}). All the images are synthesized; the image on the leftmost column is the source, while the one in the rightmost is the target synthesized image.}
\label{fig:prodpoly_linear_fashionmnist_interpolations}
\end{figure}

\begin{figure}
\centering
    \centering
    \includegraphics[width=1\linewidth]{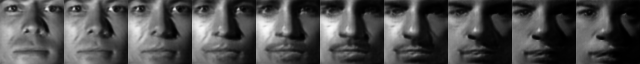} \\
    \includegraphics[width=1\linewidth]{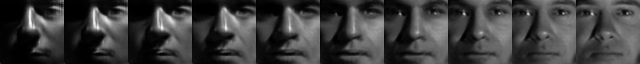} \\
    \includegraphics[width=1\linewidth]{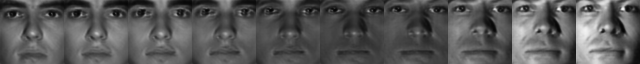} \\
    \includegraphics[width=1\linewidth]{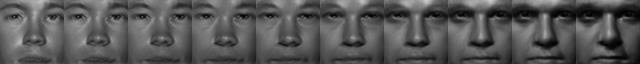} \\
    \includegraphics[width=1\linewidth]{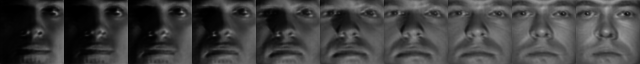} \\
    \includegraphics[width=1\linewidth]{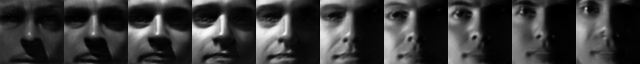} \\
    \includegraphics[width=1\linewidth]{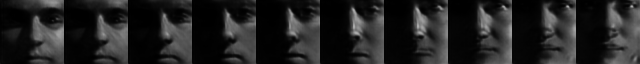}
\caption{Linear interpolation in the latent space of \modelname{} (when trained on facial images~\cite{georghiades2001few}). As in Fig.~\ref{fig:prodpoly_linear_fashionmnist_interpolations}, the generator includes only linear blocks; the image on the leftmost column is the source, while the one in the rightmost is the target image.}
\label{fig:prodpoly_linear_yaleb_interpolations}
\end{figure}

Two experiments are conducted with a polynomial generator (Fashion-Mnist and YaleB). We perform a linear interpolation in the latent space when trained with Fashion-Mnist~\cite{xiao2017fashion} and with YaleB~\cite{georghiades2001few} and visualize the results in Figs.~\ref{fig:prodpoly_linear_fashionmnist_interpolations}, \ref{fig:prodpoly_linear_yaleb_interpolations}, respectively. Note that the linear interpolation generates plausible images and navigates among different categories, e.g. trousers to sneakers or trousers to t-shirts. Equivalently, it can linearly traverse the latent space from a fully illuminated to a partly dark face.

\subsection{Linear classification}
\label{ssec:prodpoly_linear_classification}
To empirically illustrate the power of the polynomial, we use \resnet{} without activations for classification. Residual Network (\resnet)~\cite{he2016deep,srivastava2015highway} and its variants~\cite{huang2017densely, wang2018mixed, xie2017aggregated, zhang2017residual, zagoruyko2016wide} have been applied to diverse tasks including object detection and image generation~\cite{grinblat2017class,gulrajani2017improved,miyato2018spectral}. 
The core component of \resnet{} is the residual block; the $t^{th}$ residual block is expressed as $\bm{z}_{t+1} = \bm{z}_t + \bm{C} \bm{z}_t$ for input $\bm{z}_t$.

We modify each residual block to express a higher order interaction, which can be achieved with \modelthree. The output of each residual block is the input for the next residual block, which makes our \resnet{} a product of polynomials. We conduct a classification experiment with CIFAR10~\cite{krizhevsky2014cifar} ($10$ classes) and CIFAR100~\cite{cifar100} ($100$ classes). Each residual block is modified in two ways: a) all the activation functions are removed, b) it is converted into an $i^{th}$ order expansion with $i \in [2, 5]$. The second order expansion (for the $t^{th}$ residual block) is expressed as $\bm{z}_{t+1} = \bm{z}_t + \bm{C} \bm{z}_t + \Big( \bm{C} \bm{z}_t \Big) * \bm{z}_t$; higher orders are constructed similarly by performing a Hadamard product of the last term with $\bm{z}_t$ (e.g., for third order expansion it would be $\bm{z}_{t+1} = \bm{z}_t + \bm{C} \bm{z}_t + \Big( \bm{C} \bm{z}_t \Big) * \bm{z}_t + \Big( \bm{C} \bm{z}_t \Big) * \bm{z}_t * \bm{z}_t$). The following two variations are evaluated: a) a single residual block is used in each `group layer', b) two blocks are used per `group layer'. The latter variation is equivalent to \resnet18 without activations. 

\begin{figure}
\centering
    \centering
    \includegraphics[width=0.6\linewidth]{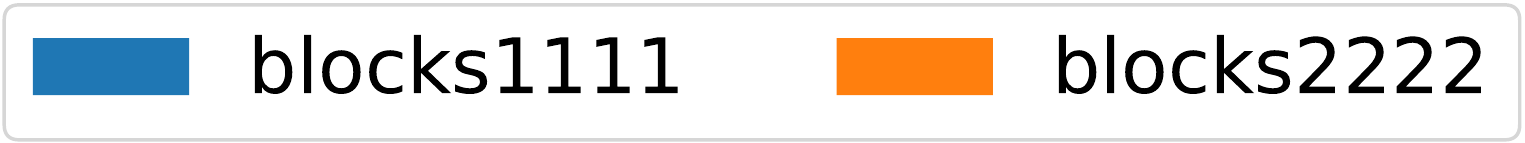} \\
    \includegraphics[width=0.48\linewidth]{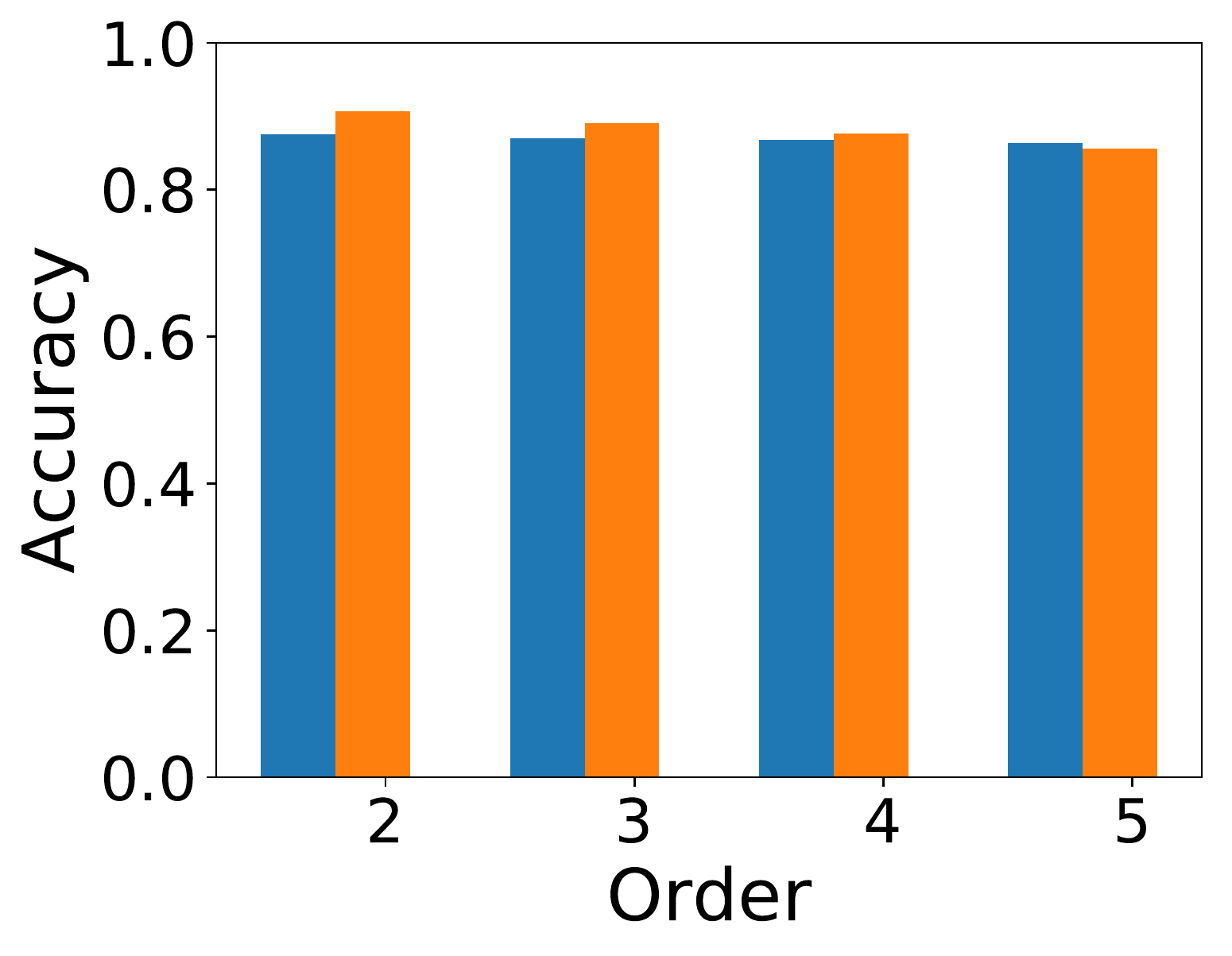}
    \includegraphics[width=0.48\linewidth]{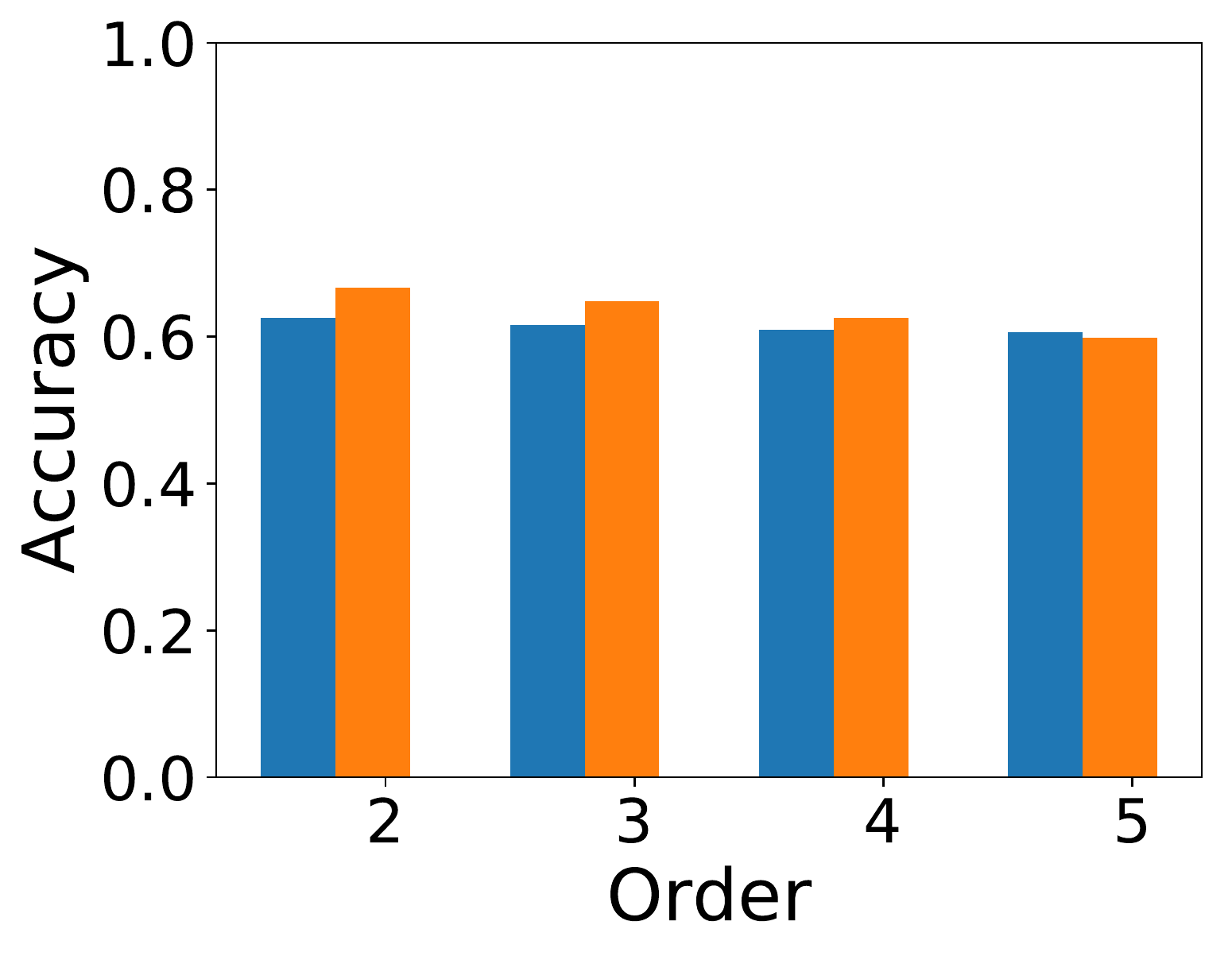}
\caption{Image classification accuracy with linear residual blocks\textsuperscript{\ref{foot:prodpoly_linear_resnet_basel}}. The schematic on the left is on CIFAR10 classification, while the one on the right is on CIFAR100 classification.}
\label{fig:prodpoly_exper_linear_resnet_accur}
\end{figure}

Each experiment is conducted $10$ times; the mean accuracy\textsuperscript{\ref{ft:prodpoly_supplementary}} is reported in Fig.~\ref{fig:prodpoly_exper_linear_resnet_accur}. We note that the same trends emerge in both datasets\footnote{\label{foot:prodpoly_linear_resnet_basel} The performance of the baselines, i.e. \resnet18 \textbf{without} activation functions, is $0.391$ and $0.168$ for CIFAR10 and CIFAR100 respectively. However, we emphasize that the original \resnet{} was not designed to work without activation functions. The performance of \resnet18 in CIFAR10 and CIFAR100 \textbf{with} activation functions is $0.945$ and $0.769$ respectively.}.
The performance remains similar irrespective of the the amount of residual blocks in the group layer. The performance is affected by the order of the expansion, i.e. higher orders cause a decrease in the accuracy. Our conjecture is that this can be partially attributed to overfitting (note that a $3^{rd}$ order expansion for the $2222$ block - in total 8 res. units - yields a polynomial of $3^8$ power), however we defer a detailed study of this in a future version of our work. Nevertheless, in all cases without activations the accuracy is close to the original \resnet18 with activation functions.

 \section{Experiments}
\label{sec:prodpoly_experiments}
We conduct three experiments against state-of-the-art models in three diverse tasks: image generation, image classification, and graph representation learning. In each case, the baseline considered is converted into an instance of our family of $\Pi$-nets and the two models are compared.

\subsection{Image generation}
\label{ssec:prodpoly_experiment_stylegan}
The robustness of \modelname{} in image generation is assessed in two different architectures/datasets below. 

\textbf{SNGAN on CIFAR10}: In the first experiment, the architecture of SNGAN~\cite{miyato2018spectral} is selected as a strong baseline on CIFAR10~\cite{krizhevsky2014cifar}. The baseline includes $3$ residual blocks in the generator and the discriminator. 

The generator is converted into a $\Pi$-net, where each residual block is a single order of the polynomial. We implement two versions, one with a single polynomial (\modeltwo) and one with product of polynomials (where each polynomial uses \modeltwo). In our implementation $\bm{A}_{[n]}$ is a thin FC layer, $(\bm{B}\matnot{n})^T \bm{b}\matnot{n}$ is a bias vector and $\bm{S}\matnot{n}$ is the transformation of the residual block. Other than the aforementioned modifications, the hyper-parameters (e.g. discriminator, learning rate, optimization details) are kept the same as in \cite{miyato2018spectral}. 

Each network has run for 10 times and the mean and variance are reported. The popular Inception Score (IS)~\citep{salimans2016improved} and the Frechet Inception Distance (FID)~\citep{heusel2017gans} are used for quantitative evaluation. Both scores extract feature representations from a pre-trained classifier (the Inception network~\citep{szegedy2015going}). 

The quantitative results are summarized in Table~\ref{tab:prodpoly_exper_cifar10_sota}. In addition to SNGAN and our two variations with polynomials, we have added the scores of \cite{grinblat2017class,gulrajani2017improved,du2019implicit, hoshen2019non,lucas2019adversarial} as reported in the respective papers. Note that the single polynomial already outperforms the baseline, while the \modelname{} boosts the performance further and achieves a substantial improvement over the original SNGAN.

\begin{table}
    \caption{IS/FID scores on CIFAR10~\citep{krizhevsky2014cifar} generation. The scores of \cite{grinblat2017class,gulrajani2017improved} are added from the respective papers as using similar residual based generators. The scores of \cite{du2019implicit, hoshen2019non, lucas2019adversarial} represent alternative generative models. \modelname outperform the compared methods in both metrics.}
     \begin{tabular}{|c | c | c|} 
     \hline
     \multicolumn{3}{|c|}{Image generation on CIFAR10}\\ 
     \hline
     Model & IS ($\uparrow$) & FID ($\downarrow$)\\
     \hline
     SNGAN & $8.06\pm 0.10$ & $19.06\pm 0.50$\\
     \hline
     \modeltwo (Sec.~\ref{ssec:prodpoly_single_poly}) & $8.30\pm 0.09$ & $17.65\pm 0.76$\\
     \hline
     \modelname & $\bm{8.49\pm 0.11}$ & $\bm{16.79\pm 0.81}$\\
     \cmidrule[\heavyrulewidth](){1 - 3}
     CSGAN-\cite{grinblat2017class} & $7.90\pm0.09$ & -\\
     \hline
     WGAN-GP-\cite{gulrajani2017improved} & $7.86\pm0.08$ & -\\
     \hline
     CQFG-\cite{lucas2019adversarial}  & $8.10$ & $18.60$\\
     \hline
     EBM~\cite{du2019implicit} & $6.78$ & $38.2$ \\
     \hline
     GLANN~\cite{hoshen2019non} & - & $46.5\pm0.20$\\
      \hline
     \end{tabular}
     \label{tab:prodpoly_exper_cifar10_sota}
\end{table}

\textbf{StyleGAN on FFHQ}: StyleGAN~\cite{karras2018style} is the state-of-the-art architecture in image generation. The generator is composed of two parts, namely: (a) the mapping network, composed of 8 FC layers, and (b) the synthesis network, which is based on ProGAN~\cite{karras2017progressive} and progressively learns to synthesize high quality images. The sampled noise is transformed by the mapping network and the resulting vector is then used for the synthesis network. As discussed in the introduction StyleGAN is already an instance of the $\Pi$-net family, due to AdaIN. 
Specifically, the $k^{th}$ AdaIN layer is $\bm{h}_k = (\bm{A}_k^T\bm{w}) * n(c(\bm{h}_{k-1}))$, where $n$ is a normalization, $c$ is a convolution and $\bm{w}$ is the transformed noise $\bm{w} = MLP(\bm{z})$ (mapping network). 
This is equivalent to our NCP model by setting $\bm{S}\matnot{k}^T$ as the convolution operator. 

In this experiment we illustrate how simple modifications, using our family of products of polynomials, further improve the representation power. We make a minimal modification in the mapping network, while fixing the rest of the hyper-parameters. In particular, we convert the mapping network into a polynomial (specifically a \modeltwo), which makes the generator a product of two polynomials.  

\begin{figure}
\centering
    \centering
    \includegraphics[width=0.245\linewidth]{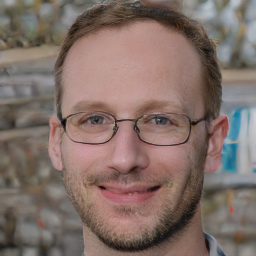}\hspace{-0.5mm}
    \includegraphics[width=0.245\linewidth]{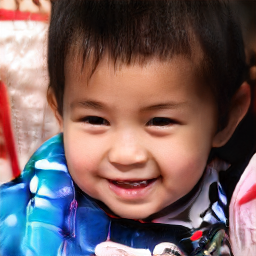}\hspace{-0.5mm}
    \includegraphics[width=0.245\linewidth]{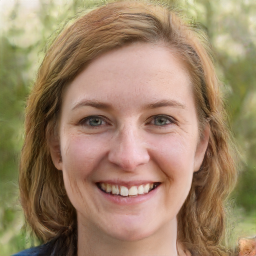}\hspace{-0.5mm}
    \includegraphics[width=0.245\linewidth]{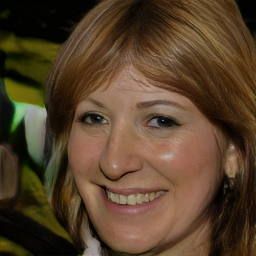}
\caption{Samples synthesized from \modelname{} (trained on FFHQ).}
\label{fig:prodpoly_stylegan_visual}
\end{figure}

The Flickr-Faces-HQ Dataset (FFHQ) dataset~\cite{karras2018style} which includes $70,000$ images of high-resolution faces is used. All the images are resized to $256\times256$. The best FID scores of the two methods (in $256\times 256$ resolution) are $\bm{6.82}$ for ours and $7.15$ for the original StyleGAN, respectively. That is, our method improves the results by $5\%$. Synthesized samples of our approach are visualized in Fig.~\ref{fig:prodpoly_stylegan_visual}.

\subsection{Classification}
We perform two experiments on classification: a) audio classification, b) image classification. 

\textbf{Audio classification}: The goal of this experiment is twofold: a) to evaluate \resnet{} on a distribution that differs from that of natural images, b) to validate whether higher-order blocks make the model more expressive. The core assumption is that we can increase the expressivity of our model, or equivalently we can use less residual blocks of higher-order to achieve performance similar to the baseline. 

The performance of \resnet{} is evaluated on the Speech Commands dataset~\cite{warden2018speech}. 
The dataset includes $60,000$ audio files; each audio contains a single word of a duration of one second. There are $35$ different words (classes) with each word having $1,500 - 4,100$ recordings. Every audio file is converted into a mel-spectrogram of resolution $32\times32$. 

The baseline is a \resnet34 architecture; we use second-order residual blocks to build the \modelres{} to match the performance of the baseline.
The quantitative results are added in Table~\ref{tab:newton_resnet_speech_command}. The two models share the same accuracy, however \modelres{} includes $38\%$ fewer parameters. This result validates our assumption that our model is more expressive and with even fewer parameters, it can achieve the same performance.

\begin{table}[h]
 \caption{Speech classification with \resnet. The accuracy of the compared methods is similar, but \modelres{} has $38\%$ fewer parameters. The symbol `\# par' abbreviates the number of parameters (in millions). }
\centering
     \begin{tabular}{|c | c | c | c|} 
         \hline
         \multicolumn{4}{|c|}{Speech Commands classification with \resnet}\\ 
         \hline
         Model & \# blocks & \# par & Accuracy\\
        \hline
         \resnet34 & $[3, 4, 6, 3]$ &  $21.3$ & $0.951 \pm 0.002$\\
         \hline
         \modelres & $[3, 3, 3, 2]$ &  $\bm{13.2}$ & $0.951 \pm 0.002$\\
         \hline
     \end{tabular}
 \label{tab:newton_resnet_speech_command}
\end{table}

\begin{table*}[h]
\caption{Image classification (ImageNet) with \resnet. ``Speed'' refers to the inference speed (images/s) of each method.}
\centering
\begin{tabular}{|c | c | c | c| c | c |}
\hline
\multicolumn{6}{|c|}{ImageNet classification with \resnet}\\ 
\hline
Model & \# Blocks  & Top-1 error ($\%$) & Top-5 error ($\%$) & Speed  & Model Size\\
\hline
\resnet50   & $[3, 4, 6, 3]$  &  23.570 & 6.838 & 8.5K & 50.26 MB  \\
\hline
\modelres50 & $[3, 4, 6, 3]$  &  22.875 & 6.358 & 7.5K & 68.81 MB \\
\hline
\end{tabular}
\label{tab:prodpoly_resnet_imagenet}
\end{table*}

\textbf{Image classification}:
We perform a large-scale classification experiment on ImageNet~\cite{russakovsky2015imagenet}. 
We choose float16 instead of float32 to achieve $3.5\times$ acceleration and reduce the GPU memory consumption by $50\%$. 
To stabilize the training, the second order of each residual block is normalized with a hyperbolic tangent unit. SGD with momentum $0.9$, weight decay $10^{-4}$ and a mini-batch size of $1024$ is used.   
The initial learning rate is set to $0.4$ and decreased by a factor of $10$ at $30, 60$, and $80$ epochs. Models are trained for $90$ epochs from scratch, using linear warm-up of the learning rate during first five epochs according to \citet{goyal2017accurate}. For other batch sizes due to the limitation of GPU memory, we linearly scale the learning rate (e.g. $0.1$ for batch size $256$). 

\begin{figure}[!h]
    \centering

    \includegraphics[width=1.1\linewidth]{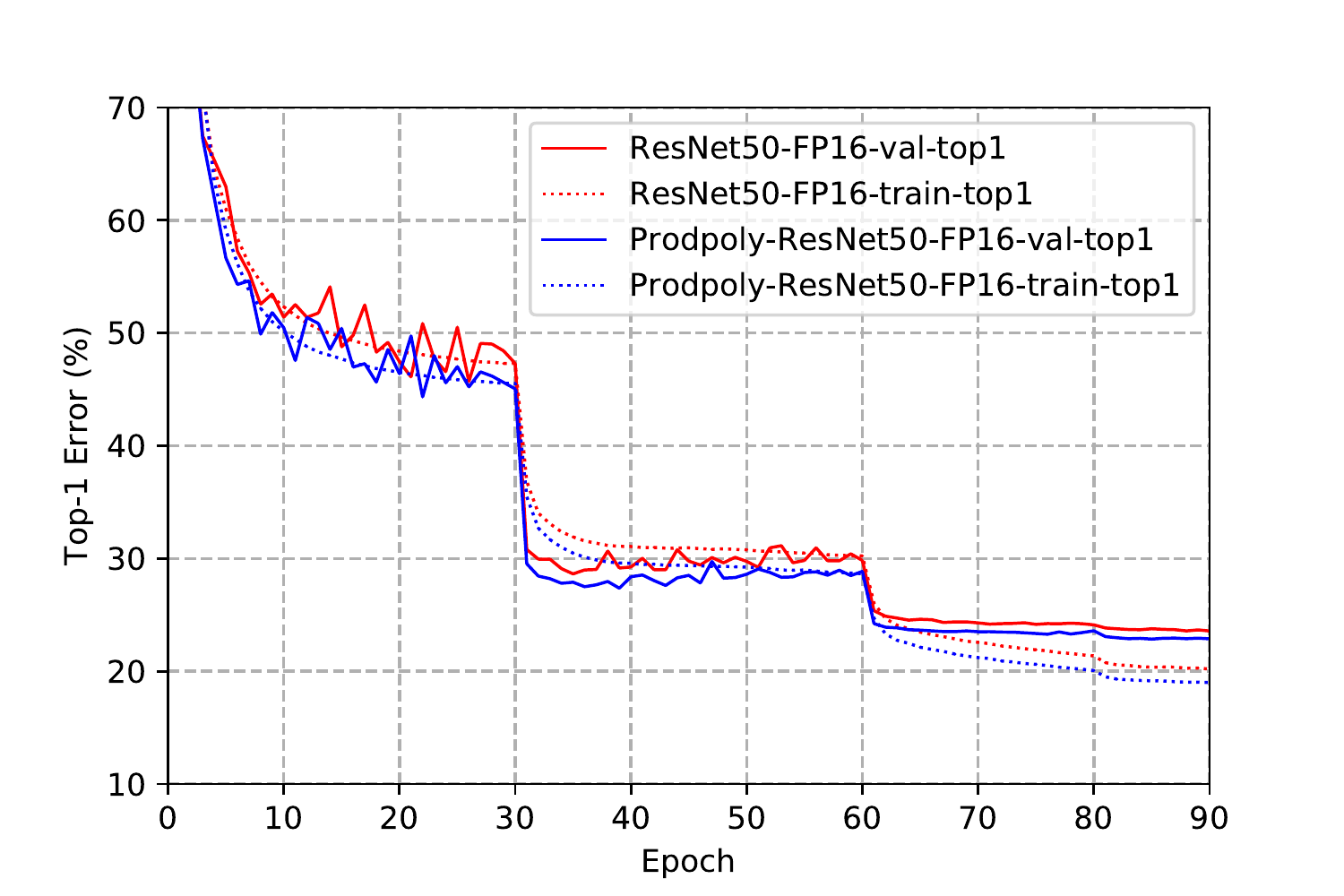}
\caption{Top-1 error on \resnet50 and \modelres50. Note that \modelres{} performs consistently better during the training; the improvement is also reflected in the validation performance.}
\label{fig:prodpoly_top1_imagenet}
\end{figure}

The Top-1 error throughout the training is visualized in Fig.~\ref{fig:prodpoly_top1_imagenet}, while the validation results are added in Table~\ref{tab:prodpoly_resnet_imagenet}. For a fair comparison, we report the results from our training in both the original \resnet{} and \modelres{}\footnote{The performance of the original \resnet~\citet{he2016deep} is inferior to the one reported here and in \citet{hu2018squeeze}.}. \modelres{} consistently improves the performance with an extremely small increase in computational complexity and model size. Remarkably, \modelres50 achieves a single-crop Top-5 validation error of $6.358\%$, exceeding \resnet50 (6.838\%) by 0.48\%. 

\begin{table}[t]
\setlength{\tabcolsep}{5.3pt}
\renewcommand{\arraystretch}{1} 
\centering
\scalebox{0.9}{
\begin{tabular}{|l||c|c|c|c|}
\hline
& error (mm) ($\downarrow$) &  speed (ms) ($\downarrow$)\\\hline
GAT \cite{velickovic2018graph} & 0.732  & 11.04\\\hline
FeastNet \cite{verma2018feastnet} & 0.623  & 6.64\\\hline
MoNet \cite{monti2017geometric} &  0.583 & 7.59 \\\hline
SpiralGNN \cite{Bouritsas_2019_ICCV} & 0.635 &  4.27\\\hline\hline
\modelname{} (simple) & 0.530 &  4.98 \\\hline
\modelname{} (simple - linear) & 0.529 & 4.79\\\hline
\modelname{} (full)  & 0.476 & 5.30\\\hline
\modelname{} (full - linear)  & \textbf{0.474} & 5.14\\\hline
\end{tabular}
}
\caption{\modelname{} vs 1st order graph learnable operators for mesh autoencoding. Note that even without using activation functions the proposed methods significantly improve upon the state-of-the-art.}
\label{prodpoly_graphs}
\end{table}

\subsection{3D Mesh representation learning}
\label{ssec:prodpoly_mesh_representation_learning_experiment}
Below, we evaluate higher order correlations in graph related tasks. We experiment with 3D deformable meshes of fixed topology~\cite{ranjan2018generating}, \ie the connectivity of the graph $\mathcal{G}= \{\mathcal{V},\mathcal{E}\}$ remains the same and each different shape is defined as a different signal $\bm{x}$ on the vertices of the graph: $\bm{x}:\mathcal{V}\to\mathbb{R}^d$. As in the previous experiments, we extend a state-of-the-art operator, namely spiral convolutions \cite{Bouritsas_2019_ICCV}, with the \modelname{} formulation and test our method on the task of autoencoding 3D shapes. We use the existing architecture and hyper-parameters of \cite{Bouritsas_2019_ICCV}, thus showing that \modelname{} can be used as a plug-and-play operator to existing models, turning the aforementioned one into a Spiral $\Pi$-Net.
Our implementation uses a product of polynomials, where each polynomial is a specific instantiation of \eqref{eq:prodpoly_model2}: $\boutvar_{n} = \Big(\bm{S}\matnot{n}^T\boutvar_{n-1}\Big) * \Big(\bm{S}\matnot{n}^T \boutvar_{n-1}\Big) + \bm{S}\matnot{n}^T\boutvar_{n-1}$, $\boutvar = \boutvar_{n} + \bm{\beta}$, where $\bm{S}$ is the spiral convolution operator written in matrix form.\footnote{Stability of the optimization is ensured by applying vertex-wise instance normalization on the 2nd order term.}
We use this model (\textit{\modelname{} simple}) to showcase how to increase the expressivity without adding new blocks in the architecture. This model can be also re-interpreted as a learnable polynomial activation function as in \cite{kileel2019expressive}. We also show the results of our complete model (\textit{\modelname{} full}), where $\bm{A}\matnot{n}$ is a different spiral convolution.

In Table \ref{prodpoly_graphs} we compare the reconstruction error of the autoencoder and the inference time of our method with the baseline spiral convolutions, as well as with the best results reported in \cite{Bouritsas_2019_ICCV} that other (more computationally involved - see inference time in table \ref{prodpoly_graphs}) graph learnable operators yielded. Interestingly, we manage to outperform all previously introduced models even when discarding the activation functions across the entire network. Thus, expressivity increases without having to increase the depth or the width of the architecture, as usually done by ML practitioners, and with small sacrifices in terms of inference time.

 \section{Discussion}
\label{sec:prodpoly_discussion}

In this work, we have introduced a new class of DCNNs, called $\Pi$-Nets that perform function approximation using a polynomial neural network. Our $\Pi$-Nets can be efficiently implemented via a special kind of skip connections that lead to high-order polynomials, naturally expressed with tensorial factors. The proposed formulation extends the standard compositional paradigm of overlaying linear operations with activation functions. We motivate our method by a sequence of experiments without activation functions that showcase the expressive power of polynomials, and demonstrate that $\Pi$-Nets are effective in both discriminative, as well as generative tasks. Trivially modifying state-of-the-art architectures in image generation, image and audio classification and mesh representation learning, the performance consistently imrpoves. In the future, we aim to explore the link between different decompositions and the resulting architectures and theoretically analyse their expressive power.

\section{Acknowledgements}
\label{sec:prodpoly_acks}

We are thankful to Nvidia for the hardware donation and Amazon web services for the cloud credits. The work of GC, SM, and GB was partially funded by an Imperial College DTA. The work of JD was partially funded by Imperial President's PhD Scholarship. The work of SZ was partially funded by the EPSRC Fellowship DEFORM: Large Scale Shape Analysis of Deformable Models of Humans (EP/S010203/1) and a Google Faculty Award. An early version with single polynomials for the generative settings can be found in \cite{chrysos2019polygan}.

{\small
\bibliographystyle{ieee_fullname}
\bibliography{egbib}

\begin{thebibliography}{10}\itemsep=-1pt

\bibitem{Bouritsas_2019_ICCV}
Giorgos Bouritsas, Sergiy Bokhnyak, Stylianos Ploumpis, Michael Bronstein, and
  Stefanos Zafeiriou.
\newblock Neural 3d morphable models: Spiral convolutional networks for 3d
  shape representation learning and generation.
\newblock In {\em International Conference on Computer Vision (ICCV)}, 2019.

\bibitem{brock2019large}
Andrew Brock, Jeff Donahue, and Karen Simonyan.
\newblock Large scale gan training for high fidelity natural image synthesis.
\newblock In {\em International Conference on Learning Representations (ICLR)},
  2019.

\bibitem{chrysos2019polygan}
Grigorios Chrysos, Stylianos Moschoglou, Yannis Panagakis, and Stefanos
  Zafeiriou.
\newblock Polygan: High-order polynomial generators.
\newblock {\em arXiv preprint arXiv:1908.06571}, 2019.

\bibitem{deng2009imagenet}
Jia Deng, Wei Dong, Richard Socher, Li-Jia Li, Kai Li, and Li Fei-Fei.
\newblock Imagenet: A large-scale hierarchical image database.
\newblock In {\em Conference on Computer Vision and Pattern Recognition
  (CVPR)}, pages 248--255, 2009.

\bibitem{denil2013predicting}
Misha Denil, Babak Shakibi, Laurent Dinh, Marc'Aurelio Ranzato, and Nando
  De~Freitas.
\newblock Predicting parameters in deep learning.
\newblock In {\em Advances in neural information processing systems (NeurIPS)},
  pages 2148--2156, 2013.

\bibitem{ding2017c}
Caiwen Ding, Siyu Liao, Yanzhi Wang, Zhe Li, Ning Liu, Youwei Zhuo, Chao Wang,
  Xuehai Qian, Yu Bai, Geng Yuan, et~al.
\newblock Circnn: accelerating and compressing deep neural networks using
  block-circulant weight matrices.
\newblock In {\em Proceedings of the 50th Annual IEEE/ACM International
  Symposium on Microarchitecture}, pages 395--408, 2017.

\bibitem{du2019implicit}
Yilun Du and Igor Mordatch.
\newblock Implicit generation and generalization in energy-based models.
\newblock In {\em Advances in neural information processing systems (NeurIPS)},
  2019.

\bibitem{frankle2018lottery}
Jonathan Frankle and Michael Carbin.
\newblock The lottery ticket hypothesis: Finding sparse, trainable neural
  networks.
\newblock In {\em International Conference on Learning Representations (ICLR)},
  2019.

\bibitem{fukushima1980neocognitron}
Kunihiko Fukushima.
\newblock Neocognitron: A self-organizing neural network model for a mechanism
  of pattern recognition unaffected by shift in position.
\newblock {\em Biological cybernetics}, 36(4):193--202, 1980.

\bibitem{georghiades2001few}
Athinodoros~S Georghiades, Peter~N Belhumeur, and David~J Kriegman.
\newblock From few to many: Illumination cone models for face recognition under
  variable lighting and pose.
\newblock {\em {IEEE} Transactions on Pattern Analysis and Machine Intelligence
  (T-PAMI)}, (6):643--660, 2001.

\bibitem{glorot2010understanding}
Xavier Glorot and Yoshua Bengio.
\newblock Understanding the difficulty of training deep feedforward neural
  networks.
\newblock In {\em International Conference on Artificial Intelligence and
  Statistics (AISTATS)}, pages 249--256, 2010.

\bibitem{goodfellow2014generative}
Ian Goodfellow, Jean Pouget-Abadie, Mehdi Mirza, Bing Xu, David Warde-Farley,
  Sherjil Ozair, Aaron Courville, and Yoshua Bengio.
\newblock Generative adversarial nets.
\newblock In {\em Advances in neural information processing systems (NeurIPS)},
  2014.

\bibitem{goyal2017accurate}
Priya Goyal, Piotr Doll{\'a}r, Ross Girshick, Pieter Noordhuis, Lukasz
  Wesolowski, Aapo Kyrola, Andrew Tulloch, Yangqing Jia, and Kaiming He.
\newblock Accurate, large minibatch sgd: Training imagenet in 1 hour.
\newblock {\em arXiv:1706.02677}, 2017.

\bibitem{grinblat2017class}
Guillermo~L Grinblat, Lucas~C Uzal, and Pablo~M Granitto.
\newblock Class-splitting generative adversarial networks.
\newblock {\em arXiv preprint arXiv:1709.07359}, 2017.

\bibitem{gulrajani2017improved}
Ishaan Gulrajani, Faruk Ahmed, Martin Arjovsky, Vincent Dumoulin, and Aaron~C
  Courville.
\newblock Improved training of wasserstein gans.
\newblock In {\em Advances in neural information processing systems (NeurIPS)},
  pages 5767--5777, 2017.

\bibitem{han2015learning}
Song Han, Jeff Pool, John Tran, and William Dally.
\newblock Learning both weights and connections for efficient neural network.
\newblock In {\em Advances in neural information processing systems (NeurIPS)},
  pages 1135--1143, 2015.

\bibitem{he2016deep}
Kaiming He, Xiangyu Zhang, Shaoqing Ren, and Jian Sun.
\newblock Deep residual learning for image recognition.
\newblock In {\em Conference on Computer Vision and Pattern Recognition
  (CVPR)}, pages 770--778, 2016.

\bibitem{heusel2017gans}
Martin Heusel, Hubert Ramsauer, Thomas Unterthiner, Bernhard Nessler, and Sepp
  Hochreiter.
\newblock Gans trained by a two time-scale update rule converge to a local nash
  equilibrium.
\newblock In {\em Advances in neural information processing systems (NeurIPS)},
  pages 6626--6637, 2017.

\bibitem{hoshen2019non}
Yedid Hoshen, Ke Li, and Jitendra Malik.
\newblock Non-adversarial image synthesis with generative latent nearest
  neighbors.
\newblock In {\em Conference on Computer Vision and Pattern Recognition
  (CVPR)}, pages 5811--5819, 2019.

\bibitem{hu2018squeeze}
Jie Hu, Li Shen, and Gang Sun.
\newblock Squeeze-and-excitation networks.
\newblock In {\em Conference on Computer Vision and Pattern Recognition
  (CVPR)}, pages 7132--7141, 2018.

\bibitem{huang2017densely}
Gao Huang, Zhuang Liu, Laurens Van Der~Maaten, and Kilian~Q Weinberger.
\newblock Densely connected convolutional networks.
\newblock In {\em Conference on Computer Vision and Pattern Recognition
  (CVPR)}, pages 4700--4708, 2017.

\bibitem{huang2017arbitrary}
Xun Huang and Serge Belongie.
\newblock Arbitrary style transfer in real-time with adaptive instance
  normalization.
\newblock In {\em International Conference on Computer Vision (ICCV)}, pages
  1501--1510, 2017.

\bibitem{ioffe2015batch}
Sergey Ioffe and Christian Szegedy.
\newblock Batch normalization: Accelerating deep network training by reducing
  internal covariate shift.
\newblock In {\em International Conference on Machine Learning (ICML)}, 2015.

\bibitem{ivakhnenko1971polynomial}
Alexey~Grigorevich Ivakhnenko.
\newblock Polynomial theory of complex systems.
\newblock {\em transactions on Systems, Man, and Cybernetics}, (4):364--378,
  1971.

\bibitem{karras2017progressive}
Tero Karras, Timo Aila, Samuli Laine, and Jaakko Lehtinen.
\newblock Progressive growing of gans for improved quality, stability, and
  variation.
\newblock In {\em International Conference on Learning Representations (ICLR)},
  2018.

\bibitem{karras2018style}
Tero Karras, Samuli Laine, and Timo Aila.
\newblock A style-based generator architecture for generative adversarial
  networks.
\newblock In {\em Conference on Computer Vision and Pattern Recognition
  (CVPR)}, 2019.

\bibitem{kileel2019expressive}
Joe Kileel, Matthew Trager, and Joan Bruna.
\newblock On the expressive power of deep polynomial neural networks.
\newblock In {\em Advances in neural information processing systems (NeurIPS)},
  2019.

\bibitem{kingma2014adam}
Diederik~P Kingma and Jimmy Ba.
\newblock Adam: A method for stochastic optimization.
\newblock In {\em International Conference on Learning Representations (ICLR)},
  2015.

\bibitem{kolda2009tensor}
Tamara~G Kolda and Brett~W Bader.
\newblock Tensor decompositions and applications.
\newblock {\em SIAM review}, 51(3):455--500, 2009.

\bibitem{cifar100}
Alex Krizhevsky, Vinod Nair, and Geoffrey Hinton.
\newblock Cifar-100 (canadian institute for advanced research).

\bibitem{krizhevsky2014cifar}
Alex Krizhevsky, Vinod Nair, and Geoffrey Hinton.
\newblock The cifar-10 dataset.
\newblock {\em online: http://www. cs. toronto. edu/kriz/cifar. html}, 55,
  2014.

\bibitem{krizhevsky2012imagenet}
Alex Krizhevsky, Ilya Sutskever, and Geoffrey~E Hinton.
\newblock Imagenet classification with deep convolutional neural networks.
\newblock In {\em Advances in neural information processing systems (NeurIPS)},
  pages 1097--1105, 2012.

\bibitem{lecun1998gradient}
Yann LeCun, L{\'e}on Bottou, Yoshua Bengio, Patrick Haffner, et~al.
\newblock Gradient-based learning applied to document recognition.
\newblock {\em Proceedings of the IEEE}, 86(11):2278--2324, 1998.

\bibitem{li2003sigma}
Chien-Kuo Li.
\newblock A sigma-pi-sigma neural network (spsnn).
\newblock {\em Neural Processing Letters}, 17(1):1--19, 2003.

\bibitem{liu2015deep}
Ziwei Liu, Ping Luo, Xiaogang Wang, and Xiaoou Tang.
\newblock Deep learning face attributes in the wild.
\newblock In {\em International Conference on Computer Vision (ICCV)}, pages
  3730--3738, 2015.

\bibitem{lucas2019adversarial}
Thomas Lucas, Konstantin Shmelkov, Karteek Alahari, Cordelia Schmid, and Jakob
  Verbeek.
\newblock Adversarial training of partially invertible variational
  autoencoders.
\newblock {\em arXiv preprint arXiv:1901.01091}, 2019.

\bibitem{miyato2018spectral}
Takeru Miyato, Toshiki Kataoka, Masanori Koyama, and Yuichi Yoshida.
\newblock Spectral normalization for generative adversarial networks.
\newblock In {\em International Conference on Learning Representations (ICLR)},
  2018.

\bibitem{monti2017geometric}
Federico Monti, Davide Boscaini, Jonathan Masci, Emanuele Rodola, Jan Svoboda,
  and Michael~M Bronstein.
\newblock Geometric deep learning on graphs and manifolds using mixture model
  cnns.
\newblock In {\em Conference on Computer Vision and Pattern Recognition
  (CVPR)}, 2017.

\bibitem{nair2010rectified}
Vinod Nair and Geoffrey~E Hinton.
\newblock Rectified linear units improve restricted boltzmann machines.
\newblock In {\em Proceedings of the 27th international conference on machine
  learning (ICML-10)}, pages 807--814, 2010.

\bibitem{nikol2013analysis}
S.M. Nikol'skii.
\newblock {\em Analysis III: Spaces of Differentiable Functions}.
\newblock Encyclopaedia of Mathematical Sciences. Springer Berlin Heidelberg,
  2013.

\bibitem{oh2003polynomial}
Sung-Kwun Oh, Witold Pedrycz, and Byoung-Jun Park.
\newblock Polynomial neural networks architecture: analysis and design.
\newblock {\em Computers \& Electrical Engineering}, 29(6):703--725, 2003.

\bibitem{park2019semantic}
Taesung Park, Ming-Yu Liu, Ting-Chun Wang, and Jun-Yan Zhu.
\newblock Semantic image synthesis with spatially-adaptive normalization.
\newblock In {\em Conference on Computer Vision and Pattern Recognition
  (CVPR)}, pages 2337--2346, 2019.

\bibitem{paszke2017automatic}
Adam Paszke, Sam Gross, Soumith Chintala, Gregory Chanan, Edward Yang, Zachary
  DeVito, Zeming Lin, Alban Desmaison, Luca Antiga, and Adam Lerer.
\newblock Automatic differentiation in {PyTorch}.
\newblock In {\em NeurIPS Workshops}, 2017.

\bibitem{ramachandran2017searching}
Prajit Ramachandran, Barret Zoph, and Quoc~V Le.
\newblock Searching for activation functions.
\newblock {\em arXiv preprint arXiv:1710.05941}, 2017.

\bibitem{ranjan2018generating}
Anurag Ranjan, Timo Bolkart, Soubhik Sanyal, and Michael~J Black.
\newblock Generating 3d faces using convolutional mesh autoencoders.
\newblock In {\em European Conference on Computer Vision (ECCV)}, pages
  704--720, 2018.

\bibitem{reddi2019convergence}
Sashank~J Reddi, Satyen Kale, and Sanjiv Kumar.
\newblock On the convergence of adam and beyond.
\newblock In {\em International Conference on Learning Representations (ICLR)},
  2018.

\bibitem{russakovsky2015imagenet}
Olga Russakovsky, Jia Deng, Hao Su, Jonathan Krause, Sanjeev Satheesh, Sean Ma,
  Zhiheng Huang, Andrej Karpathy, Aditya Khosla, Michael Bernstein, et~al.
\newblock Imagenet large scale visual recognition challenge.
\newblock {\em International Journal of Computer Vision (IJCV)},
  115(3):211--252, 2015.

\bibitem{salimans2016improved}
Tim Salimans, Ian Goodfellow, Wojciech Zaremba, Vicki Cheung, Alec Radford, and
  Xi Chen.
\newblock Improved techniques for training gans.
\newblock In {\em Advances in neural information processing systems (NeurIPS)},
  pages 2234--2242, 2016.

\bibitem{saxe2013exact}
Andrew~M Saxe, James~L McClelland, and Surya Ganguli.
\newblock Exact solutions to the nonlinear dynamics of learning in deep linear
  neural networks.
\newblock In {\em International Conference on Learning Representations (ICLR)},
  2014.

\bibitem{schmidhuber2015deep}
J{\"u}rgen Schmidhuber.
\newblock Deep learning in neural networks: An overview.
\newblock {\em Neural networks}, 61:85--117, 2015.

\bibitem{shin1991pi}
Yoan Shin and Joydeep Ghosh.
\newblock The pi-sigma network: An efficient higher-order neural network for
  pattern classification and function approximation.
\newblock In {\em International Joint Conference on Neural Networks}, volume~1,
  pages 13--18, 1991.

\bibitem{sidiropoulos2017tensor}
Nicholas~D Sidiropoulos, Lieven De~Lathauwer, Xiao Fu, Kejun Huang, Evangelos~E
  Papalexakis, and Christos Faloutsos.
\newblock Tensor decomposition for signal processing and machine learning.
\newblock {\em IEEE Transactions on Signal Processing}, 65(13):3551--3582,
  2017.

\bibitem{simonyan2014very}
Karen Simonyan and Andrew Zisserman.
\newblock Very deep convolutional networks for large-scale image recognition.
\newblock In {\em International Conference on Learning Representations (ICLR)},
  2015.

\bibitem{srivastava2015highway}
Rupesh~Kumar Srivastava, Klaus Greff, and J{\"u}rgen Schmidhuber.
\newblock Highway networks.
\newblock {\em arXiv preprint arXiv:1505.00387}, 2015.

\bibitem{stone1948generalized}
Marshall~H Stone.
\newblock The generalized weierstrass approximation theorem.
\newblock {\em Mathematics Magazine}, 21(5):237--254, 1948.

\bibitem{szegedy2015going}
Christian Szegedy, Wei Liu, Yangqing Jia, Pierre Sermanet, Scott Reed, Dragomir
  Anguelov, Dumitru Erhan, Vincent Vanhoucke, and Andrew Rabinovich.
\newblock Going deeper with convolutions.
\newblock In {\em Conference on Computer Vision and Pattern Recognition
  (CVPR)}, pages 1--9, 2015.

\bibitem{chainer_learningsys2015}
Seiya Tokui, Kenta Oono, Shohei Hido, and Justin Clayton.
\newblock Chainer: a next-generation open source framework for deep learning.
\newblock In {\em NeurIPS Workshops}, 2015.

\bibitem{ulyanov2016instance}
Dmitry Ulyanov, Andrea Vedaldi, and Victor Lempitsky.
\newblock Instance normalization: The missing ingredient for fast stylization.
\newblock {\em arXiv preprint arXiv:1607.08022}, 2016.

\bibitem{velickovic2018graph}
Petar Veli{\v{c}}kovi{\'{c}}, Guillem Cucurull, Arantxa Casanova, Adriana
  Romero, Pietro Li{\`{o}}, and Yoshua Bengio.
\newblock Graph attention networks.
\newblock {\em International Conference on Learning Representations (ICLR)},
  2018.

\bibitem{verma2018feastnet}
Nitika Verma, Edmond Boyer, and Jakob Verbeek.
\newblock Feastnet: Feature-steered graph convolutions for 3d shape analysis.
\newblock In {\em Conference on Computer Vision and Pattern Recognition
  (CVPR)}, 2018.

\bibitem{voutriaridis2003ridge}
Christodoulos Voutriaridis, Yiannis~S Boutalis, and Basil~G Mertzios.
\newblock Ridge polynomial networks in pattern recognition.
\newblock In {\em EURASIP Conference focused on Video/Image Processing and
  Multimedia Communications}, volume~2, pages 519--524, 2003.

\bibitem{wang2018mixed}
Wenhai Wang, Xiang Li, Jian Yang, and Tong Lu.
\newblock Mixed link networks.
\newblock In {\em International Joint Conferences on Artificial Intelligence
  (IJCAI)}, 2018.

\bibitem{warden2018speech}
Pete Warden.
\newblock Speech commands: A dataset for limited-vocabulary speech recognition.
\newblock {\em arXiv preprint arXiv:1804.03209}, 2018.

\bibitem{xiao2017fashion}
Han Xiao, Kashif Rasul, and Roland Vollgraf.
\newblock Fashion-mnist: a novel image dataset for benchmarking machine
  learning algorithms.
\newblock {\em arXiv preprint arXiv:1708.07747}, 2017.

\bibitem{xie2017aggregated}
Saining Xie, Ross Girshick, Piotr Doll{\'a}r, Zhuowen Tu, and Kaiming He.
\newblock Aggregated residual transformations for deep neural networks.
\newblock In {\em Proceedings of the IEEE conference on computer vision and
  pattern recognition}, pages 1492--1500, 2017.

\bibitem{xiong2007training}
Yan Xiong, Wei Wu, Xidai Kang, and Chao Zhang.
\newblock Training pi-sigma network by online gradient algorithm with penalty
  for small weight update.
\newblock {\em Neural computation}, 19(12):3356--3368, 2007.

\bibitem{yunpeng2017sharing}
Chen Yunpeng, Jin Xiaojie, Kang Bingyi, Feng Jiashi, and Yan Shuicheng.
\newblock Sharing residual units through collective tensor factorization in
  deep neural networks.
\newblock In {\em International Joint Conferences on Artificial Intelligence
  (IJCAI)}, 2018.

\bibitem{zagoruyko2016wide}
Sergey Zagoruyko and Nikos Komodakis.
\newblock Wide residual networks.
\newblock {\em arXiv preprint arXiv:1605.07146}, 2016.

\bibitem{zhang2017residual}
Ke Zhang, Miao Sun, Tony~X Han, Xingfang Yuan, Liru Guo, and Tao Liu.
\newblock Residual networks of residual networks: Multilevel residual networks.
\newblock {\em IEEE Transactions on Circuits and Systems for Video Technology},
  28(6):1303--1314, 2017.

\bibitem{zhao2017learning}
Shengjia Zhao, Jiaming Song, and Stefano Ermon.
\newblock Learning hierarchical features from deep generative models.
\newblock In {\em International Conference on Machine Learning (ICML)}, pages
  4091--4099, 2017.

\end{thebibliography}
}

\end{document}